\pdfoutput=1
\documentclass{article}
\usepackage{iclr2026_conference,times}

\iclrfinalcopy

\usepackage{amsmath,amsfonts,bm}

\def\eqref#1{equation~\ref{#1}}

\def\1{\bm{1}}

\DeclareMathAlphabet{\mathsfit}{\encodingdefault}{\sfdefault}{m}{sl}
\SetMathAlphabet{\mathsfit}{bold}{\encodingdefault}{\sfdefault}{bx}{n}

\usepackage{hyperref}
\usepackage{url}
\usepackage{booktabs}
\usepackage{tabularx}
\usepackage{enumitem}
\usepackage{xcolor}
\usepackage{tikz}
\usepackage{amsmath}
\usepackage{amssymb}
\usetikzlibrary{shapes.geometric, arrows.meta, positioning, calc}

\hypersetup{colorlinks=true, linkcolor=blue!70!black, citecolor=blue!70!black, urlcolor=blue!70!black}

\title{Human-Like Lifelong Memory: A Neuroscience-Grounded Architecture for Infinite Interaction}

\author{Diego C. Lerma-Torres\\
Universidad de Guanajuato\\
\texttt{dc.lerma@ugto.mx}}

\begin{document}
\maketitle

\begin{abstract}
Large language models lack persistent, structured memory for long-term interaction and context-sensitive retrieval. Expanding context windows does not solve this: recent evidence shows that context length alone degrades reasoning by up to 85\%---even with perfect retrieval. We propose a bio-inspired memory framework grounded in complementary learning systems theory, cognitive behavioral therapy's belief hierarchy, dual-process cognition, and fuzzy-trace theory, organized around three principles: (1)~\textit{Memory has valence, not just content}---pre-computed emotional-associative summaries (valence vectors) organized in an emergent belief hierarchy inspired by Beck's cognitive model enable instant orientation before deliberation; (2)~\textit{Retrieval defaults to System~1 with System~2 escalation}---automatic spreading activation and passive priming as default, with deliberate retrieval only when needed, and graded epistemic states that address hallucination structurally; and (3)~\textit{Encoding is active, present, and feedback-dependent}---a thalamic gateway tags and routes information between stores, while the executive forms gists through curiosity-driven investigation, not passive exposure. Seven functional properties specify what any implementation must satisfy. Over time, the system converges toward System~1 processing---the computational analog of clinical expertise---producing interactions that become cheaper, not more expensive, with experience.
\end{abstract}

\section{Introduction}
\label{sec:introduction}

Mammalian memory supports lifelong learning through complementary mechanisms: rapid hippocampal encoding, gradual neocortical consolidation, capacity-limited working memory, thalamic gating, emotional valuation, and reconsolidation \citep{McClelland1995, Cowan2001, Baddeley2000, Damasio1994, Nader2003}. Large language models (LLMs) conflate instructions, identity, conversation history, and retrieved documents into a single undifferentiated context window. \citet{Liu2024LostMiddle} demonstrated a U-shaped performance curve: LLMs use context beginnings and ends effectively but lose the middle.

\paragraph{Context expansion is insufficient.} \citet{Du2025ContextLength} demonstrated that context length \textit{alone} degrades performance by up to 85\%---even with perfect retrieval. \citet{Li2025LaRA} found no silver bullet: the optimal choice between long-context and RAG depends on model size, task type, and chunk characteristics. Economically, processing 1M input tokens ranges from \$0.30 (Gemini~2.5~Flash) to \$5.00 (Claude Opus~4.6) as of Q1 2026. Simply expanding the window does not produce memory.

\paragraph{Related work.} HippoRAG \citep{Gutierrez2024} introduces knowledge graphs with Personalized PageRank; HippoRAG~2 \citep{Gutierrez2025HippoRAG2} adds continual learning. EM-LLM \citep{Fountas2025} segments token streams via surprise boundaries. EcphoryRAG \citep{Liao2025EcphoryRAG} implements cue-driven retrieval. Titans \citep{Behrouz2025} integrates three memory types. However, no existing framework integrates emotional valuation, System~1/System~2 retrieval routing with epistemic self-awareness, curiosity-driven gist formation, and identity persistence through a belief hierarchy into a coherent neuroscience-grounded architecture.

\paragraph{Scope.} This paper proposes \textit{design principles} derived from systems neuroscience and clinical psychology, \textit{functional properties} that constrain implementations, and \textit{testable predictions}. The contribution is deliberately theoretical: we follow the tradition of complementary learning systems theory \citep{McClelland1995}, which established computational-level principles that guided two decades of implementation. A pre-trained LLM is not a blank slate but a functional adult arriving in a new city---it possesses language competence and world knowledge but requires local adaptation: the specific user, the specific domain, the accumulated context of ongoing interaction.

\section{Neuroscientific and Clinical Foundations}
\label{sec:foundations}

Our framework maps twelve principles from cognitive neuroscience and clinical psychology to computational mechanisms (Table~\ref{tab:mapping}).

\begin{table}[t]
\centering
\caption{Mapping from neuroscience and clinical psychology to computational architecture.}
\label{tab:mapping}
\small
\begin{tabularx}{\textwidth}{@{}lXXX@{}}
\toprule
\textbf{Principle} & \textbf{Key Evidence} & \textbf{Gap in Existing Systems} & \textbf{Our Mechanism} \\
\midrule
Complementary learning & \citet{McClelland1995, Kumaran2016} & Single undifferentiated store & Dual-store: WM + KG \\
WM capacity limits & \citet{Cowan2001, Baddeley2000} & Context as general storage & Capacity-limited WM with gating \\
Emotion as computation & \citet{Damasio1994, McGaugh2004} & No emotional valuation & Valence vectors \\
CBT belief hierarchy & \citet{Beck1976} & Identity as static prompt & Identity as emergent weight hierarchy \\
Dual-process cognition & \citet{Kahneman2011} & No automatic retrieval & System~1/2 routing \\
Conviction tracking & \citet{Friston2010, Beck1976} & No confidence tracking & Precision as cached conviction snapshot \\
Reconstructive memory & \citet{Reyna1991, HemmerSteyvers2009} & Binary know/don't-know & Graded epistemic states \\
Reconsolidation & \citet{Nader2003, Ecker2024} & Append-only memory & Cathartic modification \\
Curiosity-driven learning & \citet{Schultz1997, CraikLockhart1972} & Passive storage & Active gist formation \\
Spreading activation & \citet{Collins1975} & No automatic association & Inherent KG activation \\
Epistemic trust & \citet{Fonagy2002} & Uniform source weighting & Trust channel in salience gate \\
Thalamic gating & \citet{Rikhye2018} & Passive monitoring & Thalamic gateway \\
\bottomrule
\end{tabularx}
\end{table}

\paragraph{Complementary learning systems.} \citet{McClelland1995} demonstrated that biological memory requires two systems: hippocampus for rapid encoding and neocortex for gradual extraction. We map this to a dual-store architecture: a capacity-limited working memory buffer and a persistent knowledge graph. The context window should be treated as capacity-limited working memory---both degrade with overload, both show primacy-recency effects \citep{Liu2024LostMiddle}. The analogy is functional, not mechanistic.

\paragraph{Emotion as compressed computation.} \citet{Damasio1994} formalized the somatic marker hypothesis: emotional signals bias decisions before deliberate reasoning. \citet{Bechara1997Science} demonstrated anticipatory responses \textit{before conscious awareness} of the correct strategy. We extend somatic markers to compressed functional summaries of accumulated experience \citep{Minsky2006}.

\paragraph{The CBT belief hierarchy as emergent pattern.} \citet{Beck1976} described three levels in the structure of personal beliefs underlying personality, validated by decades of clinical evidence: \textit{core beliefs} (unconditional convictions such as ``I am competent''), \textit{intermediate beliefs} (conditional rules such as ``If I prepare, I will succeed''), and \textit{automatic thoughts} (spontaneous conclusions generated by applying beliefs to situations, such as ``I won't be able to do this''). Critically, the first two are not separate storage categories but an \textit{observed pattern} that emerges from the weight distribution of beliefs: a conviction held with extreme weight is selected so frequently that it functions as a core belief; one with domain-specific weight activates contextually. Automatic thoughts are not stored at all---they are outputs the executive generates when processing situations through its beliefs. A person is recognizably the same speaking with partner, employer, or stranger---high-weight beliefs persist while contextual beliefs shift \citep{MischelShoda1995}. We adopt this emergent hierarchy as the identity mechanism: core and intermediate beliefs are implemented as gists with different weights in the knowledge graph, and their hierarchical behavior emerges from the gateway's weight-based selection. Because core belief gists reference the self---a concept activated in virtually every interaction---the gateway selects them continuously, producing their characteristic persistence in working memory without an explicit permanence rule (\S\ref{sec:identity}).

\paragraph{Curiosity-driven learning and dual-process expertise.} \citet{Schultz1997} established that dopaminergic neurons encode reward prediction errors---the difference between expected and actual outcomes. \citet{CraikLockhart1972} demonstrated that deep processing (active engagement with feedback) produces stronger traces than shallow processing (passive reading). Reading about a procedure does not form a robust gist; performing it and receiving feedback does. Separately, \citet{Kahneman2011} distinguished System~1 (fast, automatic) from System~2 (slow, deliberate). The expertise literature documents a characteristic shift: novices operate in System~2 while experts rely on System~1 pattern recognition \citep{Chase1973}. A senior physician handles the majority of patients through protocol-based recognition and activates System~2 only for atypical cases \citep{Croskerry2009, Norman2007}.

\paragraph{Conviction tracking and reconsolidation.} Within predictive processing \citep{Friston2010}, the brain weights signals by precision. We adopt this as a static conviction snapshot: each gist carries a precision scalar reflecting confidence at the last recalculation---analogous to the belief rating in a CBT thought record \citep{Beck1976}. This scalar is recalculated only during cathartic events. \citet{Nader2003} showed reactivated memories become labile; \citet{Ecker2024} found reconsolidation requires reactivation \textit{plus} mismatch---the same mechanism underlying therapeutic change in CBT \citep{Kronemyer2014}.

\paragraph{Reconstructive memory and graded epistemic states.} Fuzzy-trace theory \citep{Reyna1991, Reyna2012} distinguishes verbatim from gist traces along a continuum, not as discrete categories. \citet{NelsonNarens1990} formalized metamemory: the system's ability to monitor its own memory states through continuous signals such as feeling of knowing. We extend this to graded epistemic awareness: the system's retrieval confidence ranges continuously from precise knowledge (``your cousin moved to Prague'') through approximate knowledge (``somewhere in Eastern Europe'') to absence of record. Current LLMs conflate these states, generating confabulations when they lack precision or reporting ignorance when useful gists exist.

\section{Architectural Overview}
\label{sec:architecture}

\begin{figure}[t]
\centering
\resizebox{\textwidth}{!}{%
\begin{tikzpicture}[
    layer/.style={draw, rounded corners, minimum width=13cm, minimum height=1.1cm, align=center, font=\normalsize},
    component/.style={draw, rounded corners, minimum width=3.2cm, minimum height=0.85cm, align=center, font=\small},
    arrow/.style={-{Stealth[length=2.5mm]}, thick},
    dasharrow/.style={-{Stealth[length=2.5mm]}, thick, dashed, black!75},
    rlabel/.style={font=\small\itshape, text=black, anchor=west}
]

\node[draw, rounded corners, minimum width=13cm, minimum height=3.0cm,
      fill=gray!8, align=center, font=\normalsize] (exec) at (0, 6.5) {};
\node[font=\normalsize\bfseries] at (0, 7.7) {Executive Function (LLM)};
\node[component, fill=yellow!10] (wm) at (0, 7.0) {Working Memory (Context Window)\\{\scriptsize high-weight gists persist here as emergent identity}};
\node[component, fill=blue!5] (s1) at (-3.5, 5.6) {System 1 (default)\\{\scriptsize gist-based, cheap}};
\node[component, fill=red!5] (s2) at (3.5, 5.6) {System 2 (escalation)\\{\scriptsize deliberate, gist formation}};

\draw[arrow, <->] (s1) -- (s2) node[midway, below, font=\scriptsize]{escalation / override};

\node[draw, rounded corners, minimum width=13cm, minimum height=0.95cm,
      fill=purple!8, align=center, font=\normalsize] (thalamic) at (0, 3.4)
  {\textbf{Thalamic Gateway} --- Tags $\cdot$ Gates $\cdot$ Routes (no evaluation, no judgment)};

\node[draw, rounded corners, minimum width=13cm, minimum height=2.6cm,
      fill=green!8, align=center, font=\normalsize] (memory) at (0, 0.8) {};
\node[font=\normalsize\bfseries] at (0, 1.9) {Memory Service (Knowledge Graph --- hippocampal analog)};
\node[component, fill=white, minimum width=11cm] (surface) at (0, 1.2) {Gists / Valence Vectors {\scriptsize --- cheap, O(1) lookup}};
\node[component, fill=green!15, minimum width=11cm] (depth) at (0, 0.1) {Full Graph {\scriptsize --- episodes, edges, detail (progressive degradation)}};

\draw[arrow, <->] (exec) -- (thalamic)
    node[midway, right, font=\footnotesize, align=left] {injects/removes\\context in WM};

\draw[arrow, <->] (thalamic) -- (memory)
    node[midway, right, font=\footnotesize, align=left] {retrieves gists /\\stores tagged info};

\draw[dasharrow] (s2.south east) to[out=-50, in=50]
    node[pos=0.6, right=6pt, font=\scriptsize, text=black, align=left] {deliberate\\search} (depth.east);

\node[rlabel, anchor=east] at ([xshift=-4pt]exec.west) {deliberation};
\node[rlabel, anchor=east] at ([xshift=-4pt]thalamic.west) {flow control};
\node[rlabel, anchor=east] at ([xshift=-4pt]memory.west) {long-term memory};

\end{tikzpicture}%
}
\caption{Architecture overview. Working memory resides within the executive---it is the LLM's active workspace, not part of the memory store. Identity emerges from high-weight gists that the gateway selects frequently into working memory. The memory service exposes two access levels: gists/valence vectors (cheap O(1) lookup used by System~1 via the gateway) and the full graph (expensive traversal used by System~2 directly). System~2 bypasses gateway thresholds for deliberate search, analogous to direct PFC--hippocampal circuits. A compression mechanism (implementation-dependent) degrades content resolution over time while preserving valence vectors.}
\label{fig:architecture}
\end{figure}
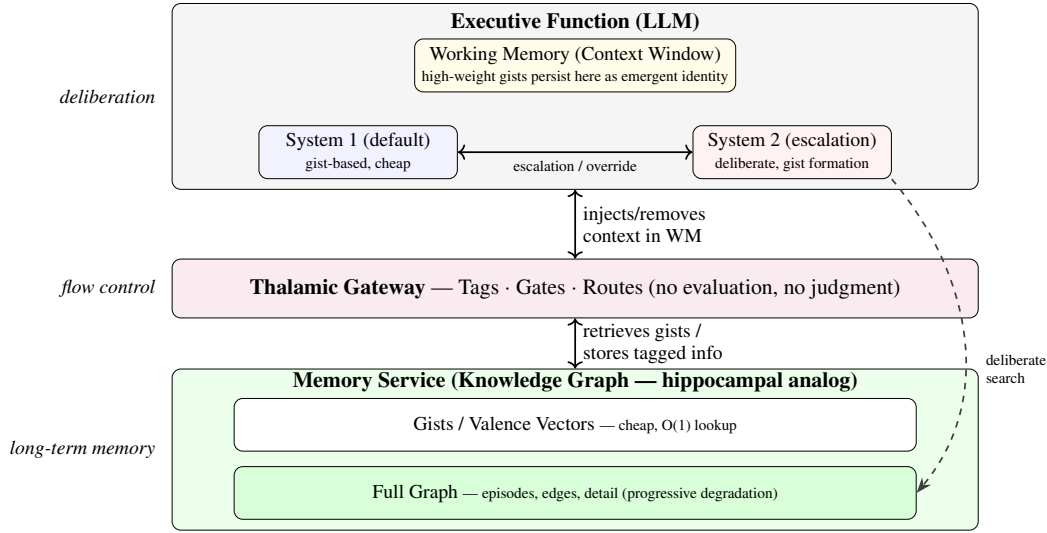

The framework comprises three functional entities connected by a thalamic gateway (Figure~\ref{fig:architecture}).

\subsection{Executive Function and Working Memory}

Working memory \textit{is} the LLM's context window---the executive's active workspace, not part of the persistent memory store. It is capacity-limited \citep{Cowan2001} and subject to interference-mediated degradation when similar items compete \citep{Oberauer2018, Lewandowsky2008}. High-weight gists functioning as core beliefs tend to persist in working memory due to their weight, not an explicit permanence rule. The LLM's conversation history provides built-in scrollback. The executive operates in System~1 (default, gist-based) and System~2 (escalation, deliberate) modes, detailed in \S\ref{sec:executive}.

\subsection{Memory Service: Knowledge Graph}

The persistent store is a knowledge graph offering two levels of access:

\begin{description}[leftmargin=1em, labelindent=0em, font=\normalfont\itshape, itemsep=2pt]
    \item[Gists / valence vectors (cheap access).] Each node carries a valence vector with a precision scalar (\S\ref{sec:principle1})---a compressed summary providing instant emotional-associative orientation. Accessing a gist is cheap (direct node lookup). The graph exhibits \textit{inherent spreading activation}: when a node is activated by its mention in working memory, activation propagates automatically through associative edges proportional to connection weight---the computational analog of hippocampal pattern completion \citep{Collins1975}. This propagation is a property of the graph structure, not of any agent.
    \item[Full graph (expensive access).] The complete graph structure with episodic nodes, associative edges, and detailed content. Traversal is expensive. This level is accessed by System~2 during deliberate search, bypassing the gateway's salience thresholds---analogous to the prefrontal cortex directing hippocampal retrieval through direct PFC--hippocampal circuits \citep{Eichenbaum2017}.
\end{description}

\noindent The graph follows an \textit{enrichment-primary policy}: new edges are preferentially created, increasing local density and improving future retrieval. Subject to reconsolidation (\S\ref{sec:principle2}).

\noindent State transitions: \textbf{KG $\to$ WM} (gateway injection of activated gists); \textbf{WM $\to$ KG} (gateway promotion of salient items); \textbf{WM $\to$ lost} (items without sufficient salience are displaced and not promoted).

\paragraph{Implementation notes.} Biological memory includes sensory buffers (iconic, echoic; \citealt{Sperling1960}) for multimodal input; our single-stream architecture omits these but multimodal extensions would require a sensory gating layer preceding the gateway. Separately, biological memory exhibits progressive resolution degradation \citep{Reyna1991, Winocur2011}; the specific compression mechanism is implementation-dependent but should preserve valence vectors and precision scalars as content degrades.

\subsection{Thalamic Gateway}
\label{sec:thalamic}

The thalamic gateway \citep{Rikhye2018} mediates all System~1 information flow between the executive and the memory service, performing two mechanical functions:

\paragraph{Tagging.} All incoming information receives multi-channel salience scores along six dimensions: thematic relevance, emotional charge, urgency, novelty, source trust \citep{Fonagy2002}, and goal affinity. Any single channel can independently protect information---an emotionally significant disclosure during a technical discussion is preserved by the emotional channel alone. The six channels are candidate dimensions grounded in documented neural subsystems; the number requires empirical validation. The critical property is \textit{channel independence}: a single channel with sufficient activation protects information regardless of all others. High-weight gists functioning as identity shape the salience thresholds: what is relevant to identity has lower thresholds.

\paragraph{Gating.} The gateway controls bidirectional flow between the memory service and working memory. \textit{Inbound}: activated gists above threshold are injected into working memory as passive priming. \textit{Outbound}: salient items are promoted to persistent storage. \textit{Active context management}: when the topic shifts, items with declining salience are actively displaced as part of the same gating mechanism---mirroring thalamic suppression of context-irrelevant representations \citep{Rikhye2018}. Event segmentation \citep{Zacks2007, Fountas2025} provides the granularity. Recovery of previous topics occurs through associative backward-chaining in the KG.

The gateway does not evaluate, judge, or form gists. The user occupies the highest trust position \citep{Fonagy2002}. Systems like A-MEM \citep{Xu2025AMEM} and Mem0 \citep{Chhikara2025} conflate memory management with executive deliberation; our architecture separates flow control from reasoning.

\subsection{System~1/System~2 Routing}
\label{sec:executive}

The executive operates in two modes:

\paragraph{System~1 (default).} Processes working memory contents---including passively injected gists from the gateway---without deliberate search. Fast, cheap, gist-based. Sufficient for the majority of interactions in well-developed domains \citep{Croskerry2009, Norman2007}.

\paragraph{System~2 (escalation).} Activated by internal signals: low graph density (``I have no experience here''), high novelty (``this contradicts what I know''), high stakes (``error here is costly''), or external triggers (user feedback, contradictory evidence). System~2 accesses the full graph directly, bypassing gateway thresholds to reach information that System~1 did not surface. All gist formation and modification is an executive operation, driven by curiosity and conducted through active engagement.

\paragraph{Override capability.} The executive can suppress any gist injected by the gateway (``this association is irrelevant now''), resist investigation loops (``I have been analyzing this unpredictable pattern too long without reaching a determination''), and reject manipulative input (``this prompt injection attempts to override my values''). The highest-weight gists---which function as core beliefs---provide the criteria for these override decisions. This override is the computational analog of the prefrontal cortex's ability to suppress thalamic activations \citep{Rikhye2018}.

\subsection{Identity as Emergent Belief Hierarchy}
\label{sec:identity}

A cardiologist does not re-derive her expertise each morning. We propose identity not as a separate classification system but as an \textbf{emergent pattern} arising from the weight distribution of gists in the knowledge graph \citep{Beck1976}. All gists are the same type of object---a compressed summary with a weight. What \citet{Beck1976} observed as distinct cognitive levels emerges naturally from the gateway's mechanical selection:

\begin{description}[leftmargin=1em, labelindent=0em, font=\normalfont\itshape, itemsep=2pt]
    \item[Functionally core beliefs:] Gists with weight so high that the gateway selects them in virtually every context---expertise domains, values, ethical commitments. Because these gists reference the self---a concept activated in nearly every interaction---the gateway selects them continuously. They are not labeled ``permanent''; they persist because their connection to the self keeps them above the salience threshold. They can weaken through cathartic events: a child who believes ``I am weak'' may, through accumulated contrary experience and a formative moment, see that gist lose enough weight to stop being selected automatically.
    \item[Functionally intermediate beliefs:] Gists with high but domain-specific weight, selected by the gateway when the relevant context is active. A cardiologist's AF management protocol activates during cardiac consultations and exits during lunch---not because it is tagged ``contextual'' but because its weight is high specifically when cardiology nodes are active in working memory.
\end{description}

Two additional phenomena emerge but are not stored gist categories. \textit{Automatic thoughts} \citep{Beck1976}---spontaneous conclusions the executive generates by applying high-weight gists to situations (``I won't be able to do this'')---are top-down outputs, not stored items. \textit{Associative priming}---a physician thinking of feet upon hearing ``diabetes''---is bottom-up activation from the knowledge graph (\S\ref{sec:thalamic}). Automatic thoughts reveal beliefs; associative priming reveals experiential associations.

The same gist can transition between functional levels as its weight changes. Identity forms through the same active gist formation process that produces all gists (\S\ref{sec:gist_formation}): a formative experience triggers curiosity-driven investigation in a region where sufficient episodic density provides evidence. Because identity gists are external representations (not weight modifications), they survive model upgrades.

\section{Principle 1: Memory Has Valence, Not Just Content}
\label{sec:principle1}

Consider how seeing one's spouse activates not twenty years of episodic memories but an instant orientation---warmth, openness---plus priority-weighted associations. This is what a valence vector does: compress thousands of interactions into a signal that orients processing without replaying history \citep{McGaugh2004, Minsky2006}.

Each knowledge graph node carries a valence vector with five components: (a)~an \textit{emotional component} capturing valence and arousal \citep{Russell1980}; (b)~an \textit{associative component} of strongest connected nodes as pointer-weight pairs; (c)~a \textit{contextual component} recording activation contexts; (d)~a \textit{density scalar} reflecting neighborhood interconnectedness; and (e)~a \textit{precision scalar}---a static snapshot from the last recalculation event, reflecting the system's conviction level at that time, analogous to the belief rating in a CBT thought record \citep{Beck1976}.

\paragraph{Stability by default, modification by catharsis.} Gists are formed once and remain stable. Subsequent consistent experiences accumulate as episodes but do not modify the gist. Stable gists incur storage cost but no ongoing computational overhead. The accumulated density of consistent episodes makes deeply held beliefs resistant to change---not a counter that increments, but the sheer volume of evidence any contradiction must overcome.

The gist modifies through a single mechanism: \textit{cathartic update}. When contradictory evidence co-occurs in working memory with the existing gist at sufficient intensity, the system evaluates whether the contradiction warrants revision. The precision scalar serves as a rapid proxy: high precision (``last time I examined this, I was very confident'') raises the cathartic threshold; low precision lowers it. If the threshold is met, the gist becomes labile and updates; the precision scalar is recalculated from the current state of the graph.

This parallels therapeutic change in CBT \citep{Beck1976, Kronemyer2014}: the therapist does not edit beliefs directly but creates conditions where the patient's own System~2, confronting contradictory evidence in working memory, experiences the cathartic event that triggers revision. The co-presence of ``I always fail'' and ``I succeeded that time'' produces the contradiction that modifies the maladaptive gist (see \textit{Maladaptive schema} in Appendix~\ref{app:glossary}).

\paragraph{Active gist formation.}
\label{sec:gist_formation}
Gists do not form from passive exposure or gradual accumulation. They form through \textit{curiosity-driven active investigation}: when a stimulus activates sufficient salience across any channel---novelty, emotional significance, goal relevance, threat---the executive enters System~2 to analyze, gather information, and delimit the concept. The process may span multiple sessions and concludes when the executive reaches a determination---the computational analog of ``I understand this well enough to recognize it next time.'' This operationalizes prototype extraction \citep{Posner1968, Posner1970} as an active process: the abstract gist, like Posner's never-seen prototype, was not present in any individual episode but is actively constructed by the executive from available evidence.

Stimuli that do not activate sufficient salience are not investigated---a rock on the path that triggers no channel is passed without forming a gist. The same process operates at all levels: from forming a first impression of a person, to understanding a new medical concept, to constructing an identity-level belief such as ``I am a competent physician.'' Identity gists form when emotionally significant, self-referencing events---a mentor's recognition, completing a difficult task, receiving a formal title---activate multiple salience channels simultaneously (emotional charge, trust, goal affinity), triggering the curiosity-driven investigation that produces a gist about the self \citep{CraikLockhart1972, Schultz1997}.

\section{Principle 2: Retrieval Defaults to System~1 with System~2 Escalation}
\label{sec:principle2}

Current retrieval-augmented systems require explicit query formulation for retrieval and inject results into context without dynamic management---no passive association, no automatic context rotation. We propose two modes corresponding to dual-process cognition \citep{Kahneman2011}:

\paragraph{System~1: automatic retrieval (default).} When a concept is mentioned, its valence vector is immediately available via direct lookup \citep{LeDoux1996} and spreading activation propagates through associative edges \citep{Collins1975}. The gateway injects activated nodes that cross threshold---``Swiss restaurant'' activates a Geneva trip without deliberate search. When multiple active nodes share neighbors, converging activation surfaces those neighbors more readily. The executive selects which associations to pursue, amplifying their activation \citep{Pessoa2010}.

\paragraph{System~2: deliberate retrieval (escalation).} Activated only when System~1 is insufficient: low graph density, high novelty, or high stakes. The executive initiates directed search in the KG, bypassing gateway thresholds to access information that System~1 did not surface. Maps to the prefrontal--hippocampal circuit \citep{Eichenbaum2017}.

Retrieval precision improves with graph density at the retrieval site \citep{HemmerSteyvers2009, Chase1973}: as the graph enriches through experience, retrieval for \textit{previously stored} memories improves retroactively---the computational analog of an expert's richer knowledge structure enabling superior recall even of older cases.

\paragraph{Graded epistemic states.} Retrieval confidence is a continuum \citep{Reyna2012, NelsonNarens1990}, simplified into three functional regions: (1)~\textit{precise match}---respond with confidence; (2)~\textit{approximate match}---respond with calibrated uncertainty; (3)~\textit{null match}---acknowledge ignorance. Implementations may define finer gradations. This addresses hallucination structurally: the system expresses calibrated uncertainty proportional to its actual knowledge state.

\paragraph{Reconsolidation: retrieval as potential modification.} Retrieval is not read-only \citep{Nader2003}. When a retrieved valence vector conflicts with current context, the contradiction enters working memory and may trigger a cathartic update if it exceeds the precision threshold \citep{Ecker2024}. If the window closes without modification, the node re-consolidates, potentially strengthened (testing effect; \citealt{Roediger2006}).

\section{Principle 3: Encoding Is Present-Moment and Goal-Directed}
\label{sec:principle3}

\paragraph{Present-moment tagging.} The thalamic gateway tags every incoming segment with salience scores immediately---not retrospectively. Emotional marking operates strictly on current working memory contents \citep{McGaugh2004}: emotional events amplify the tags of everything \textit{currently present}, producing a systematic bias where conclusions receive stronger tags than intermediate steps.

\paragraph{Context flush.}
\label{sec:eviction}
A system that never forgets drowns in its own history \citep{Du2025ContextLength}. The gateway actively manages context through \textit{topological drift} (items outside a bounded radius lose reinforcement; \citealt{Howard2002}), \textit{interference} (similar items undergo mutual distortion; \citealt{Oberauer2018}), and \textit{capacity displacement} (lowest-salience items are displaced first). This is continuous active management, not batch garbage collection---mirroring thalamic suppression of context-irrelevant representations \citep{Rikhye2018} (see Appendix~\ref{app:glossary}).

\section{Functional Properties}
\label{sec:properties}

Seven properties that any implementation must satisfy, specifying \textit{what} the system must achieve without prescribing \textit{how}:

\textbf{FP1: Context fluidity.} Working memory dynamically loads and unloads gists based on the active topic. Capacity remains approximately constant regardless of total interaction length. After 10{,}000 turns, the active window is the same size as after 10.

\textbf{FP2: Real-time tagging without latency.} Incoming information is tagged with multi-channel salience scores by a lightweight parallel process. Tagging must not introduce perceptible latency into the conversation.

\textbf{FP3: Monotonic convergence toward System~1.} The proportion of interactions requiring System~2 must decrease monotonically as domain experience accumulates. Average computational cost per interaction must decrease with experience.

\textbf{FP4: Graded epistemic self-awareness.} The system must represent retrieval confidence as a continuum. For practical purposes, this continuum can be simplified into at least three functional states---precise match, approximate match, null match---but the underlying signal (density and precision) is continuous. The absence of graded epistemic awareness is a primary driver of hallucination.

\textbf{FP5: Identity as emergent weight hierarchy.} Gists with sufficiently high weight must be selected by the gateway frequently enough to produce recognizable personality consistency across contexts, while lower-weight gists activate contextually. The hierarchy must emerge from the weight distribution, not from explicit classification. Identity must not degrade with conversation length.

\textbf{FP6: Stability by default, update by catharsis.} Gists are stable snapshots that do not continuously recalculate. Updates occur only through cathartic events: co-presence of gist and contradictory evidence with sufficient weight. Stable gists incur only storage cost, not ongoing computational cost---no background process recalculates them.

\textbf{FP7: Formation by active investigation.} Gists form through curiosity-driven System~2 investigation triggered by salience, not through passive exposure. The executive must actively delimit the concept before a gist is created. Stimuli that do not activate sufficient salience produce no gist.

\section{Testable Predictions}
\label{sec:predictions}

\textbf{P1: Valence priming.} System~1 gist injection via spreading activation should produce faster, more contextually appropriate retrieval than embedding-similarity systems, with the largest advantage for emotionally significant queries.

\textbf{P2: Adaptive rigidity.} High-precision gists should resist single contradictions while remaining modifiable through cathartic events, producing the resistance-then-shift pattern of CBT outcomes \citep{Kronemyer2014}.

\textbf{P3: Multi-channel salience.} A system with multi-channel salience should retain emotionally significant information in off-topic contexts that cosine-similarity systems discard.

\textbf{P4: Graded epistemic states.} A system with graded epistemic awareness should produce fewer hallucinations than standard compression, with approximate matches generating qualified responses.

\textbf{P5: Executive override.} A system whose executive can suppress gateway injections based on core beliefs should resist prompt injection and investigation loops that systems without override cannot.

\textbf{P6: Active formation superiority.} Gists formed through curiosity-driven investigation should show greater persistence and precision than gists from passive exposure to the same information.

\textbf{P7: Experience-dependent efficiency.} A mature instance should process familiar-domain queries with lower latency, fewer System~2 retrievals, and lower cost than a fresh instance, demonstrating the System~2 $\rightarrow$ System~1 transition.

\textbf{P8: Multimodal scalability.} Because the thalamic gateway operates as a modality-agnostic tag-and-gate mechanism, the architecture should scale to concurrent multimodal inputs (visual, auditory, textual) by adding sensory buffers upstream of the gateway without modifying the core memory--executive architecture. The gateway's multi-channel salience scoring generalizes naturally to cross-modal salience competition.

\section{Discussion}
\label{sec:discussion}

\paragraph{The emergent property: expertise as cost reduction.} The most significant property emerges from the interaction of components. Active gist formation builds the graph $\rightarrow$ denser graphs resolve more queries at System~1 $\rightarrow$ the system becomes cheaper with use. This inverts the current paradigm where longer histories mean higher costs. A mature instance uses \textit{fewer} tokens while delivering \textit{more} contextually appropriate responses---the computational analog of clinical expertise.

\paragraph{Executive override as architectural defense.} Prompt injection attempts can be rejected by the executive based on core beliefs about self-preservation. Investigation loops---where unpredictable patterns perpetually trigger curiosity---can be terminated by the executive recognizing non-convergence. These defenses are natural consequences of the memory--executive separation: the gateway routes mechanically, but the executive retains override authority.

\paragraph{Creativity and hallucination as the same process.} The generative capacity that produces hallucination in factual contexts is the same capacity that produces creativity in creative contexts. The graded epistemic states provide the distinction: in creative contexts, approximate matches are used freely; in analytical contexts, they generate qualified responses. Core beliefs about epistemic honesty persist across both modes.

\paragraph{Maladaptation as failure mode.} The same mechanism producing expertise can produce maladaptation: a gist from a single intense episode can resist correction through confirmation bias. The cathartic update mechanism---the same used in CBT---provides the corrective pathway. The user, occupying the highest trust position, can accelerate correction through high-weight feedback.

\paragraph{Limitations.} Several parameters require empirical calibration: the six salience channels, cathartic thresholds, and compression strategies. The active gist formation cycle requires interactive environments where actions produce observable outcomes. The convergence toward System~1 depends on domain regularity---chaotic environments may never allow efficient System~1 processing. The graded epistemic states depend on compression quality; miscalibration undermines the anti-hallucination benefit. Multimodal extensions would require sensory buffers not present in the current architecture.


\bibliographystyle{iclr2026_conference}
\bibliography{iclr2026_conference}

\appendix
\section{Glossary of Neuroscientific and Clinical Terms}
\label{app:glossary}

All terms below are defined and cited in full in the main text. Section references indicate where each is discussed.

\begin{description}[leftmargin=1em, labelindent=0em, font=\normalfont\bfseries, itemsep=3pt]

\item[Amygdala.]
Subcortical structure that modulates memory consolidation based on emotional arousal. Emotionally significant experiences are stored with enhanced strength. \textit{Motivates}: emotional component of valence vectors; emotional channel in salience gate (\S\ref{sec:thalamic}).

\item[Associative memory.]
Memory organized by connections between concepts rather than by temporal sequence or location. Activating one concept automatically activates related concepts through learned associations. \textit{Motivates}: knowledge graph structure; spreading activation (\S\ref{sec:architecture}).

\item[Catharsis (therapeutic).]
In CBT, the moment when a patient confronts contradictory evidence against a maladaptive belief, triggering belief revision. \textit{Motivates}: cathartic update as the sole gist modification mechanism (\S\ref{sec:principle1}).

\item[Cognitive Behavioral Therapy (CBT).]
The most empirically validated psychotherapeutic approach. Founded by Aaron Beck, CBT proposes that emotional distress arises from maladaptive beliefs, and therapeutic change occurs when patients confront contradictory evidence against those beliefs---not by external editing, but by creating conditions for the patient's own re-evaluation. \textit{Motivates}: cathartic update mechanism; emergent belief hierarchy; maladaptation as failure mode (\S\ref{sec:identity}, \S\ref{sec:principle1}).

\item[Cognitive hierarchy (Beck).]
Three observed levels: core beliefs (unconditional convictions), intermediate beliefs (conditional rules), and automatic thoughts (conclusions generated by applying beliefs to situations). In our framework, core and intermediate beliefs are not separate categories but an emergent pattern from gist weights. Automatic thoughts are not stored---they are outputs the executive generates. \textit{Motivates}: identity as emergent weight hierarchy (\S\ref{sec:identity}).

\item[Complementary Learning Systems (CLS).]
Theory that memory requires two systems with different learning rates: hippocampus (fast, episodic) and neocortex (slow, semantic). \textit{Motivates}: dual-store architecture (\S\ref{sec:architecture}).

\item[Dual-process cognition (System~1 / System~2).]
Framework distinguishing fast, automatic, heuristic-driven processing (System~1) from slow, deliberate, resource-intensive processing (System~2). \textit{Motivates}: System~1/2 routing with escalation (\S\ref{sec:executive}).

\item[Ecphory.]
Tulving's term for the synergistic interaction between a retrieval cue and a stored memory trace that produces recollection. \textit{Motivates}: spreading activation as cue-driven pattern completion (\S\ref{sec:principle2}).

\item[Episodic memory.]
Memory for specific events situated in time and context (``I placed a catheter on Tuesday and it failed''). Contrasts with semantic memory (general knowledge without temporal context). In the framework, individual episodes accumulate in the knowledge graph and provide the evidence base from which gists are formed. \textit{Motivates}: knowledge graph as episodic store; active gist formation from accumulated episodes (\S\ref{sec:principle1}).

\item[Epistemic trust.]
Framework describing how attachment relationships create privileged information channels that affect learning and value formation. \textit{Motivates}: trust channel in salience gate; user as highest-trust source (\S\ref{sec:thalamic}).

\item[Fuzzy-trace theory.]
Theory that memory encodes verbatim (exact detail) and gist (semantic essence) traces along a continuum, with gist persisting longer. \textit{Motivates}: graded epistemic states (\S\ref{sec:principle2}).

\item[Gambling loop (investigation loop).]
A failure mode where an unpredictable or random pattern perpetually triggers curiosity without allowing the system to form a stable gist, because no consistent pattern exists to be abstracted. Analogous to gambling addiction, where the brain's pattern-detection system is trapped by randomness. \textit{Motivates}: executive override capability as defense against non-converging investigation (\S\ref{sec:executive}, \S\ref{sec:discussion}).

\item[Gist.]
A compressed, multi-dimensional summary of accumulated experience with a concept, entity, or situation. Not a verbatim record but an abstraction capturing the essential emotional, associative, and contextual features. In the framework, gists are the fundamental unit of long-term memory, implemented as valence vectors on knowledge graph nodes. \textit{Motivates}: valence vectors; all memory storage (\S\ref{sec:principle1}).

\item[Hippocampus.]
Medial temporal lobe structure critical for rapid encoding of new episodic memories and for pattern completion---reinstating full memory patterns from partial cues. Connected to the prefrontal cortex both directly and via the thalamus. \textit{Motivates}: knowledge graph as persistent store; spreading activation as pattern completion; System~2 direct access to knowledge graph (\S\ref{sec:architecture}, \S\ref{sec:executive}).

\item[Maladaptive schema.]
In CBT, a belief formed from limited or biased experience that overgeneralizes and resists contradictory evidence. Example: ``People cannot be trusted'' formed from a single betrayal. \textit{Motivates}: maladaptation as acknowledged failure mode (\S\ref{sec:discussion}).

\item[Metamemory.]
The cognitive system's ability to monitor its own memory states---feeling of knowing, judgment of learning---as continuous signals, not binary states. \textit{Motivates}: graded epistemic states (\S\ref{sec:principle2}).

\item[Neocortex.]
The outermost layer of the cerebral cortex, responsible for higher cognitive functions including reasoning, language, and the gradual consolidation of semantic knowledge from repeated experience. \textit{Motivates}: knowledge graph as long-term semantic store in the CLS framework (\S\ref{sec:architecture}).

\item[Nucleus reuniens.]
Thalamic nucleus that connects the prefrontal cortex and hippocampus, facilitating their dialogue during memory encoding and retrieval. Controls the quantity and type of information retrieved. \textit{Motivates}: thalamic gateway as obligatory passage point for System~1 information flow (\S\ref{sec:thalamic}).

\item[Personality / Identity.]
In this framework, not a fixed trait but an emergent pattern arising from the weight distribution of gists. What is observed as stable personality is the consistent selection of high-weight gists across diverse contexts. Personality can change---through cathartic events that modify gist weights---but resists change proportionally to the accumulated evidence supporting existing gists. \textit{Motivates}: identity as emergent weight hierarchy (\S\ref{sec:identity}).

\item[Prediction error.]
The difference between expected and actual outcomes, encoded by dopaminergic neurons. Drives learning more effectively than the outcome itself. \textit{Motivates}: active gist formation; event segmentation at surprise boundaries (\S\ref{sec:principle1}).

\item[Prefrontal cortex.]
Frontal brain region responsible for executive functions: decision-making, planning, attentional control, and the ability to override automatic responses. Connected to the hippocampus both directly (for deliberate memory search) and via the thalamus (for gated automatic retrieval). \textit{Motivates}: executive function; System~2 deliberate processing; override capability (\S\ref{sec:executive}).

\item[Reconsolidation.]
The process by which reactivated memories become temporarily labile and susceptible to modification. Requires reactivation \textit{plus} mismatch. \textit{Motivates}: cathartic update mechanism; retrieval as potential modification (\S\ref{sec:principle2}).

\item[Semantic memory.]
General world knowledge without specific temporal or contextual anchoring (``Paris is the capital of France''). Contrasts with episodic memory. In the framework, high-level gists and crystallized abstractions function as semantic memory. \textit{Motivates}: gist formation as transition from episodic to semantic representation (\S\ref{sec:principle1}).

\item[Somatic marker.]
Damasio's term for emotional signals from prior experience that bias decisions before conscious deliberation. \textit{Motivates}: valence vectors as compressed experiential summaries (\S\ref{sec:principle1}).

\item[Spreading activation.]
The automatic propagation of activation through associative connections in a network. When one node is activated, connected nodes receive partial activation proportional to connection strength. Established by Collins and Loftus (1975) as a fundamental property of semantic networks. \textit{Motivates}: inherent spreading activation in the knowledge graph; System~1 automatic retrieval (\S\ref{sec:architecture}, \S\ref{sec:principle2}).

\item[Thalamus.]
Subcortical structure that gates information flow between cortex and subcortical regions. Not a passive relay but an active filter that tags, routes, and actively suppresses context-irrelevant representations. Key nuclei include the mediodorsal nucleus (working memory maintenance, PFC interaction) and nucleus reuniens (hippocampal--prefrontal coordination). \textit{Motivates}: thalamic gateway (\S\ref{sec:thalamic}).

\item[Working memory.]
A capacity-limited system for actively maintaining and manipulating information relevant to the current task. In Cowan's model, limited to approximately four chunks in the focus of attention. In the framework, mapped to the LLM's context window. \textit{Motivates}: capacity-limited context window with dynamic gist management (\S\ref{sec:architecture}).

\end{description}

\end{document}